\documentclass[letterpaper,10 pt,conference]{ieeeconf}   
\IEEEoverridecommandlockouts                              

\usepackage{graphics} 
\usepackage{epsfig} 
\usepackage{mathptmx} 
\usepackage{times} 

\usepackage{amsthm}

\usepackage{amsmath} 
\usepackage{amssymb}  
\usepackage{float}
\usepackage{cite}
\newcommand{\Et}[1]{\hat{\mathbb{E}}_t\left[#1\right]}

\usepackage[utf8]{inputenc} 
\usepackage[T1]{fontenc}    
\usepackage{hyperref}       
\usepackage{url}            
\usepackage{booktabs}       
\usepackage{amsfonts}       
\usepackage{nicefrac}       
\usepackage{microtype}      
\usepackage{mathtools}

\usepackage{amsfonts}
\usepackage{algorithm,algpseudocode}
\usepackage{wrapfig}

\newtheorem{proposition}{Proposition}

\theoremstyle{definition}

\newtheorem{definition}{Definition}

\newcommand{\lie}{\mathsf{L}_{f}}
\newcommand{\dlie}{\mathsf{L}_{f,\Delta t}}

\newcommand{\dliephi}{\mathsf{L}_{f_{\pi_{\phi}},\Delta t}}

\title{\LARGE \bf Stabilizing Neural Control Using Self-Learned Almost Lyapunov Critics}
\author{Ya-Chien Chang \and Sicun Gao
\thanks{\scriptsize This material is based upon work supported by the United States Air Force and DARPA under Contract No. FA8750-18-C-0092, AFOSR YIP FA9550-19-1-0041, NSF Career CCF 2047034, and NSF NRI 1830399.}
\thanks{\scriptsize Y.-C. Chang, S. Gao are with the Department of Computer Science and Engineering, UC San Diego, USA. (Email: yac021@eng.ucsd.edu; sicung@eng.ucsd.edu).} }

\begin{document}
\maketitle
\thispagestyle{empty}
\pagestyle{empty}

\begin{abstract}
The lack of stability guarantee restricts the practical use of learning-based methods in core control problems in robotics. We develop new methods for learning neural control policies and neural Lyapunov critic functions in the model-free reinforcement learning (RL) setting. We use sample-based approaches and the Almost Lyapunov function conditions to estimate the region of attraction and invariance properties through the learned Lyapunov critic functions. The methods enhance stability of neural controllers for various nonlinear systems including automobile and quadrotor control. 
\end{abstract}

\section{Introduction}

Ensuring stability of neural control policies is critical for the practical use of learning-based methods for control design in robotics. There has been exciting progress towards introducing control-theoretic approaches for enhancing stability in reinforcement learning and imitation learning, such as using Lyapunov methods~\cite{DBLP:conf/nips/BerkenkampTS017,DBLP:journals/jmlr/PerkinsB02,DBLP:conf/nips/ChowNDG18,DBLP:journals/corr/abs-1901-10031,DBLP:journals/corr/abs-2004-14288,DBLP:conf/cdc/JinL18,DBLP:conf/nips/ChangRG19,gallieri2019safe}, control barrier functions~\cite{DBLP:conf/aaai/ChengOMB19,taylor2019learning,DBLP:conf/aaai/AlshiekhBEKNT18}, or control-theoretic regularization~\cite{DBLP:conf/icml/ChengVOCYB19,7961277,pmlr-v120-liu20a}. However, three major questions are still open. First, while giving direct control-theoretic prior or guidance can improve performance, to what extent can a learning agent {\em self-supervise} to achieve certifiable stability? Second, stability requirements should be inherent to a system and not affected by the choice of reward functions, which often do not focus on stability. Can we achieve stability without restricting the flexibility of reward engineering in reinforcement learning? Third, the typical formulation of stability in learning is in the form of reducing expected violation of certain constraints asymptotically, which does not easily translate to certifiable behaviors when the learned policies are practically deployed. Is it possible to obtain stronger stability claims in the standard control-theoretic sense, such as for estimating region of attraction and forward invariance properties? Our goal is to positively answer these questions to improve the reliability and usability of learning-based methods for practical control problems in robotics. 

We propose new methods for incorporating Lyapunov methods in deep reinforcement learning. We design a model-free policy optimization process in which the agent attempts to formulate a Lyapunov-like function through {\em self-supervision}, in the form of a special critic function that is then used for improving stability of the learned policy, without using or being affected by the environment reward. Our work follows the framework of actor-critic Lyapunov methods recently proposed in~\cite{DBLP:journals/corr/abs-2004-14288} and extends it in the following ways. We allow the agent to self-learn the Lyapunov critic function by minimizing the Lyapunov risk~\cite{DBLP:conf/nips/ChangRG19} over its experience buffer without accessing the rewards. This Lyapunov critic function is represented as a generic feed-forward neural network, randomly initialized. This design differs from the typical choice of using a positive definite neural network to construct the Lyapunov candidate, to capture values learned from appropriately designed cost functions that only reflect stability objectives, as in~\cite{DBLP:journals/corr/abs-2004-14288,DBLP:conf/corl/RichardsB018}. The lack of restriction and guidance turns out to be beneficial. In our approach, the learning agent can formulate Lyapunov landscapes that better enforce stability, compared to existing methods. In fact, this form of self-learned Lyapunov candidate can often be shown to satisfy the Almost Lyapunov conditions~\cite{LIU2020108758}, which allows us to use sample-based analysis to estimate its region of attraction and forward invariance properties. We show that the new design is important for learning certifiably stable control policies for practical control problems such as automobile path-tracking and quadrotor control. 

Our work builds on the recent progress of model-free and sample-based Lyapunov methods~\cite{LIU2020108758,sample-lyapunov,DBLP:journals/corr/abs-2004-14288,DBLP:conf/corl/RichardsB018}. Such methods allow us to use sampled Lie derivatives of candidate Lyapunov functions to estimate stability properties of neural control policies without using analytic forms of the system dynamics. We exploit the expressiveness of neural networks for capturing Lyapunov landscapes that are too complex for conventional choices such as sum-of-squares polynomials. We believe the proposed methods further advance the promising direction of developing rigorous neural control and certification methods in reinforcement learning. 

In all, we make the following contributions. We propose new methods for training neural Lyapunov critic functions (Section IV.A) and use it to improve stability properties of neural control policies (Section IV.B) in a model-free policy optimization setting. We show that the learned Lyapunov critic function can be analyzed through sample-based methods based on the Almost Lyapunov conditions, for estimating its region of attraction and invariance properties (Section IV.C). We demonstrate the benefits of the proposed methods in comparison with standard policy optimization and existing Lyapunov-based actor-critic methods (Section V).

\section{Related Work} 
Besides the most closely related work~\cite{DBLP:journals/corr/abs-2004-14288} discussed above, our work is connected to various recent progress in safe reinforcement learning, Lyapunov methods in reinforcement learning, and data-driven methods for the analysis of control and dynamical systems. 
Safe reinforcement learning is now a large and active field~\cite{DBLP:journals/jmlr/GarciaF15,DBLP:conf/icml/AchiamHTA17} focusing on learning with various forms of soft or hard safety constraints, often formulated as Constrained Markov Decision Processes (CMDP)~\cite{Altman99constrainedmarkov}. Lyapunov methods have seen various applications in this context. It was first introduced in RL by the work of \cite{DBLP:journals/jmlr/PerkinsB02}, which uses predefined controllers and Lyapunov-based methods for learning a switching policy with safety and performance guarantees. \cite{DBLP:conf/nips/BerkenkampTS017} proposed a model-based RL framework that uses Lyapunov functions to guide safe exploration and uses Gaussian Process models of the dynamics to obtain high-performance control policies with provable stability certificates. \cite{DBLP:conf/nips/ChowNDG18} developed methods for constructing Lyapunov function in the tabular setting that can guarantee global safety of a behavior policy during training. The approach is extended to policy optimization for the continuous control setting in \cite{DBLP:journals/corr/abs-1901-10031}, showing benefits in various high-dimensional control tasks. The work in~\cite{DBLP:conf/cdc/JinL18} formulates the state-action value function for safety costs as candidate Lyapunov functions and model its derivative with Gaussian Processes with statistical guarantees on the control performance. Similarly in   \cite{DBLP:journals/corr/abs-2004-14288}, candidate Lyapunov functions are constructed from value functions with benefit in stabilizing the control performance. In the work of \cite{DBLP:conf/corl/RichardsB018,DBLP:conf/nips/ChangRG19}, neural networks are used to learn Lyapunov functions for establishing certificates for stability and safety properties.

Many other control-theoretic methods have also been proposed to improve reinforcement learning~\cite{DBLP:conf/aaai/ChengOMB19,taylor2019learning,DBLP:conf/aaai/AlshiekhBEKNT18,DBLP:conf/icml/ChengVOCYB19,7961277,pmlr-v120-liu20a,gallieri2019safe}. These methods typically involve introducing strong control-theoretic priors, or use the neural policies as an oracle to extract low-variance policies in simpler hypothesis classes. For instance, the work in \cite{DBLP:conf/aaai/ChengOMB19} uses control barrier functions to ensure safety of learned control policies. The work in \cite{DBLP:conf/icml/ChengVOCYB19} proves stability properties throughout learning by taking advantage of the robustness of control-theoretic priors. Our goal is to turn control-theoretic methods into general self-supervision methods for enhancing stability. 

\section{Preliminaries}
\begin{definition}[Dynamical Systems]
\label{def:dyn}
An $n$-dimensional controlled dynamical system is defined by 
\begin{equation}
\label{eqn:dynamics}
    \dot{x}(t)=f(x(t), u(t)), \;u(t)=g(x(t)), \; x(0)=x_{0},
\end{equation}
where $f:D\rightarrow \mathbb{R}^{n}$ is a Lipschitz-continuous vector field, $g:D\rightarrow \mathbb{R}^m$ is a control function, and $D\subseteq \mathbb{R}^{n}$ with $0,x_0\in D$ defines the state space of the system. Each $x(t) \in D$ is called a state vector and $u(t
)\in \mathbb{R}^m$ is a control vector. 
\end{definition}
\begin{definition}[Stability]
We say that the system of $(1)$ is stable at the origin if for any $\varepsilon\in\mathbb{R}^+$, there exists $\delta(\varepsilon) \in\mathbb{R}^+$ such that if $\|x(0)\|<\delta$ then $\|x(t)\|<\varepsilon$ for all $t\geq 0$. The system is asymptotically stable at the origin if it is stable and also $\lim_{t\rightarrow\infty}x(t)=0$ for all $\|x(0)\|<\delta$. \end{definition}

\begin{definition}[Lie Derivatives]\label{def:lie} 
Consider the system in (\ref{eqn:dynamics}) and let $V:D\rightarrow \mathbb{R}$ be a continuously differentiable function. The Lie derivative of $V$ over $f$ is defined as \begin{equation}
    \lie V(x)=\sum_{i=1}^{n} \frac{\partial V}{\partial x_{i}}\frac{\mathrm{d} x_i}{\mathrm{d} t}= \sum_{i=1}^{n} \frac{\partial V}{\partial x_{i}}\dot{x}_i(t).
\end{equation}
It measures the rate of change of $V$ over time along the direction of the system dynamics of $x(t)$. 
\end{definition}

\begin{definition}[Lyapunov Conditions for Asymptotic Stability]\label{def:lya}
Consider a controlled system $(1)$ with an equilibrium at the origin, i.e., $\exists \ u \in \mathbb{R}^m$ s.t. $f(0,u)=0$. Suppose there exists a continuously differentiable function $V:D\rightarrow \mathbb{R}$ satisfying $V\left(0\right)= 0$, and $\forall x \in D\setminus\{0\},  V\left({x}\right)> 0$, and $\lie V\left({x}\right)<0$. Then $V$ is a Lyapunov function. The system $f$ is asymptotically stable at the origin if such Lyapunov function $V$ can be found. 
\end{definition}
\noindent The recent work~\cite{LIU2020108758} proposed an approximate notion of Lyapunov methods named {\em Almost Lyapunov functions}, which allows the Lyapunov conditions to be violated in restricted subsets of the space while still ensuring stability properties. It is the basis of our sample-based approach for validating learned Lyapunov candidates. We will discuss this notion and our approach in detail in Section IV.C. 

We will use the standard notations for reinforcement learning in Markov Decision Processes (MDP) with state space $S$, action space $A$ and transition model $P\colon S \times S \times A \rightarrow [0,1]$. A reward function defines the reward for taking action $a$ in state $s$ and transitioning into $s'$. Let $\pi_\phi$ denote a stochastic policy parameterized by $\phi$. 
The goal of the learning agent is to maximize the expected $\gamma$-discounted cumulative return $J(\phi) = \mathbb{E}_{s_0, a_0, \ldots}\left[\sum_{t=0}^{\infty} \gamma^t r(s_t, a_t, s_{t+1})\right]$. Policy optimization methods~\cite{reinforce,a2c,trpo,acktr} estimate policy gradient and use stochastic gradient ascent to directly improve policy performance. A standard gradient estimator is 
\begin{equation}
\label{gestimate}
\hat{g} = \Et{\nabla_\phi \log \pi_\phi(a_t|s_t) \hat{A}_t}, 
\end{equation}
where $\pi_\phi$ is a stochastic policy and $\hat{A}_t$ estimates the advantage that represents the difference between the Q value of an action compared with the expected value of a state, to indicate whether an action should be taken more frequently in the future. The gradient steps will move the distribution over actions in the right direction accordingly. The expectation $\Et{\ldots}$ is estimated by the empirical average over finite batch of samples. The proximal policy optimization algorithm (PPO) \cite{ppo} applies clipping to the objective function to remove incentives for the policy to change dramatically, using:
\begin{equation}
J^{\rm CLIP}(\phi) =\Et{\min \left(r_{t}(\phi)  \hat{A}_t, {\rm clip} (r_{t}(\phi),1-\epsilon, 1+\epsilon)\hat{A}_t \right)},
\end{equation} where $r_{t}(\phi) = \pi_\phi(a_t|s_t)/\pi_\textit{old}(a_t|s_t)$ and $\epsilon$ is a hyperparameter. The clipping ensures the gradient steps do not overshoot in the policy parameter space in each policy update.  

\section{Policy Optimization with Lyapunov Critics}

We now describe how to use self-learned candidate Lyapunov functions to improve stability of neural control policies. We will name these candidate Lyapunov functions as {\em Lyapunov critics}, following~\cite{DBLP:journals/corr/abs-2004-14288}. We will first describe how to learn Lyapunov critics through sampled trajectories, and then how to integrate this critic values in advantage estimation for policy optimization. We describe how sampling-based certification of stability using Lyapunov critics, following the framework of Almost Lyapunov functions. The full procedure is as shown in Algorithm 1. The overall loop is close to standard PPO~\cite{ppo,schulman2018highdimensional} with only additional steps at Line 11-12, for learning the Lyapunov critic, and Line 17, for policy optimization guided by the new critic. We will explain these steps in the following sections. 

\begin{algorithm}[h]
    \caption{Policy Optimization with Self-Learned Almost Lyapunov Critics (POLYC)}\label{alg:polyc}
    \begin{algorithmic}[1]
         \State Initialize policy (actor) network $\pi_{\phi}$ and the reward value function network $V^r_{\eta}$
         \State Initialize Lyapunov function network $V_{\theta}$ randomly
         \State Initialize replay buffer $B$ as empty set
         \For{episodes $= 1,\dots,K$} 
             \For{$t = 1,\dots,T$}
                 \State \small Sample $a_t \sim \pi_{\phi}(a_t|s_t)$
                 \State \small Sample $s_{t+1} \sim P(s_t+1|s_t,a_t)$ 
                 \State \small $B \leftarrow B \cup {(s_t, a_t, r_t, s_{t+1})}$  
             \EndFor

                 \State Sample mini-batches of size $N$ from $B$
                 \State Compute Lyapunov risk $R_{f,N,\rho,\Delta t}(\theta)$ under $\pi_{\phi_{new}}$
                 \State $\theta \leftarrow \theta - \alpha_{\theta} \nabla_\theta R_{f,N,\rho,\Delta t}(\theta)$
             \For {each policy optimization step}
                 \State Sample mini-batches of transitions from $B$
                 \State \small $\delta_t \gets r_t + \gamma V^r_{\eta}(s_{t+1})-V^r_{\eta}(s_t)$ \Comment{cf. \cite{schulman2018highdimensional}}
                 \State \small $\hat{A}(s_t,a_t) \gets 
                 \delta_t+\gamma \delta_{t+1}+\dots+ \gamma^{T-t+1}\delta_{T-1}$
                 \State {\small $\hat A^L_{\beta}\gets\beta\min(0,-\mathsf{L}_{f_{\pi_{\phi}},\Delta t}V_{\theta}(s_t))+(1-\beta)\hat{A}(s_t,a_t)$}
                 \State
                 \small$r(\phi)\gets\pi_{\phi_{new}}(a|s)/ \pi_{\phi}(a|s)$
                 \State {\footnotesize $J^{\rm CLIP}(\phi,\beta) \gets \hat{\mathbb{E}}\left[\min\left(r(\phi)\hat A^L_{\beta}, {\rm clip}(r(\phi), 1-\epsilon, 1+\epsilon)\hat A^L_{\beta}\right)\right]$}
                 \State \small $\phi \leftarrow \phi + \alpha_{\phi} \nabla_\phi J^{\rm CLIP}(\phi,\beta)$ 
                 \State \footnotesize $\eta\gets \eta-\alpha_{\eta}\nabla_{\eta}\hat{\mathbb{E}}[(V^r_{\eta}(s_t)-G(s_t))^2]$ \Comment $G(s_t)=\sum^{\infty}_{k=0}\gamma^{k}r_{t+k}$
            \EndFor     
         \EndFor
\end{algorithmic}
\end{algorithm}

\subsection{Self-Learning Lyapunov Critics}
We represent candidate Lyapunov functions using neural networks, draw samples of the system states, and use gradient descent to learn network parameters that maximize the satisfaction of the Lyapunov conditions in Definition~\ref{def:lya}. Following~\cite{DBLP:conf/nips/ChangRG19}, such learning can be achieved by minimizing the following loss function named as the Lyapunov risk:
\begin{definition}[Empirical Lyapunov Risk~\cite{DBLP:conf/nips/ChangRG19}]\label{def:risk}
Consider dynamical system $f$ with domain $D$. Let $V_{\theta}:D\rightarrow \mathbb{R}$ be a continuously differentiable function parameterized by $\theta$. The empirical Lyapunov risk of $V_{\theta}$ over $f$ with a sampling distribution $\rho(D)$, written as $R_{f,N,\rho}\left(\theta\right)$, is defined as:
{\small\begin{eqnarray}\label{risk}
\frac{1}{N}\sum^{N}_{i=1}\bigg(\max(-V_{\theta}(s_{i}),0)+\max(0,\lie V_{\theta}(s_{i}))\bigg)+V_{\theta}^{2}(0)
\end{eqnarray}}where $s_1, ..., s_N$ are states sampled according to the sampling distribution $\rho$. The empirical Lyapunov risk is nonnegative, and when $V_{\theta}$ is a true Lyapunov function for $f$, this empirical risk attains its global minimum $R_{f,N,\rho}\left(\theta\right)=0$, regardless of the choice of the sample size $N$ and distribution $\rho$. 
\end{definition}
In this original formulation, evaluating the Lyapunov risk requires the full knowledge of the system dynamics $f$ for computing $\lie V$. Our current setting is different.  The learning agent does not know the system dynamics $f$ and can only approximate the Lie derivative $\lie V$ along sampled trajectories of the system, through finite differences:
\begin{eqnarray}\label{eq:lie}
\dlie{{V}(s)}=\frac{1}{\Delta t}\Big(V(s')-V(s)\Big),
\end{eqnarray}
where $s$ and $s'$ are two consecutive states and $\Delta t$ is the time difference between them and $\lim_{\Delta t\rightarrow 0}\dlie V(s)=\lie V(s)$. We define the corresponding discretized Lyapunov risk objective $R_{f,N,\rho,\Delta t}$ where the only change from (\ref{risk}) above is that the Lie derivative $\lie V_{\theta}(s)$ is replaced by the sampled estimate $\dlie V(s)$ for all states. 

At each iteration of the policy update (the main loop from Line 4 to 23 in Algorithm 1), we collect trajectories of samples from the controlled system $f_{\pi}$, and perform stochastic gradient descent to minimize the discretized Lyapunov risk $R_{f,N,\rho,\Delta t}(\theta)$ to learn the Lyapunov critic function $V_{\theta}$ (Line 11-12). Note that the dependency of the dynamics $f$ on the behavior policy $\pi_{\phi}$ is important. When the policy is updated, the controlled system changes its dynamics, and the candidate Lyapunov function should be learned correspondingly. Thus, we need to be able to sample from the buffer of previous trajectories, and estimate their Lie derivative values using the new policies (Line 10-11). 
This dependency makes our approach inherently on-policy in the current formulation, in the sense that the critic is always learned from the behavior policy $\pi_{\phi}$ and thus not very sample-efficient. Off-policy learning of the Lyapunov critic is possible through importance sampling and keeping track of the policy and Lyapunov critic updates, which is a promising direction that we leave open.

\subsection{Stabilization via Policy Optimization}

Once the Lyapunov critic is formulated, we use the learned candidate Lyapunov functions as an additional critic value for policy optimization. For a temporarily fixed candidate Lyapunov function $V_{\theta}$, the only term in the Lyapunov risk that is affected by policy change is the Lie derivative term $\dlie V_{\theta}$. Thus, in policy optimization we now impose the additional goal of ensuring $\dlie V(\theta)<0$ to encourage the policy update to improve the Lyapunov landscape for stabilization. We combine the standard advantage estimate $\hat{A}(s_t,a_t)$~\cite{ppo} and a clipped Lie derivative term in each policy optimization step:
\begin{eqnarray}\label{advantage}
\hat A^L_{\beta}(s_t,a_t)=(1-\beta)\hat{A}(s_t,a_t)+\beta\min(0,-\dliephi V(s_t))
\end{eqnarray}
where $\beta\in [0,1]$ balances the weights. With policy gradient methods on this advantage estimate, the second term penalizes actions that produce a positive Lie derivative which makes $\min(0,-\dliephi V(s_t))<0$. When the Lie derivative is negative, it does not bias the advantage estimate. We emphasize that although the Lie derivative term is dependent on $\pi_{\phi}$, it does not have a functional form but only accessed through sampled values based on Equation (\ref{eq:lie}). The true dynamics of the system remains unknown to the learner. 

With the definition of the advantage with Lyapunov critic in (\ref{advantage}), we can easily replace the standard advantage estimators in various on-policy algorithms. For instance, the standard policy gradient estimator becomes $\mathbb{E}_{\pi_{\phi}}[\nabla_{\phi}\log \pi_{\phi}(a_t|s_t)\hat{A}_\beta^t(s_t,a_t)]$. In Algorithm 1 we use the PPO version~\cite{ppo} of policy update (Line 19-20), which is what we use in the experiments. Optionally, we can update the $\beta$ parameter as a Lagrange multiplier, by also taking gradient steps on $\beta$ at some learning rate $\alpha$ using $\beta\gets \beta-\alpha\mathbb{E}[\dliephi V(s)]$, clipped between $[1,0]$. We have not observed much performance difference in doing so, and have excluded this step in Algorithm 1 for simplicity.   

\subsection{Validating Almost Lyapunov Conditions}

A major benefit of the proposed approach is that the self-learned Lyapunov critic allows us to estimate the region of attraction of the controlled system when learning is successful. Since we do not have access to the dynamics of the system and can only estimate the Lyapunov risk at sampled states, we can not expect to certify that the learned Lyapunov critic functions are true Lyapunov functions in the standard sense. However, recent progress in relaxed conditions for Lyapunov methods enabled the use of sample-based analysis to find region of stability. In particular, the Lyapunov critic functions can be analyzed through sampling using the Almost Lyapunov conditions~\cite{LIU2020108758} defined as follows:

\begin{proposition}[Almost Lyapunov Conditions~\cite{LIU2020108758}]\label{prop:lyp}
Consider a dynamical system in (1) defined by $f$ with domain $D\subseteq \mathbb{R}^n$ and a continuously differentiable positive definite function $V: D\rightarrow [0,\infty)$. Let $c_1,c_2>0$ be two constants and define $B$ as the region between two sublevel sets $B=\{x\in D: c_1\leq V(x)\leq c_2\}$ for $V$. Let $\Omega\subseteq B$ be a measurable set. Suppose for some $a>0$, $\max_{x\in B}\lie V(x)<a\min_{x\in B} V$, and $\forall x\in \Omega, \lie V(x)\geq -aV(x)$ and $\forall x\in \Omega\setminus B, \lie V(x)<-a V(x)$. Then there exists $\hat\varepsilon>0$ such that for any $\varepsilon\in (0,\hat\varepsilon)$, if the volume of each connected component $\Omega^*$ of $\Omega$ satisfies $\mathsf{vol}(\Omega^*)\leq \varepsilon$, then there exists $T>0$ such that for any $x_0\in D$ with $V(x_0)<c_2-r(\varepsilon)$, $x(t)$ stays within $B$ for all $t>0$ and  moreover, it converges to the sublevel set $V(x)\leq c_1+r(\varepsilon)$ for any $t>T$. Here $r(\varepsilon)=h\varepsilon^{1/n}+g\varepsilon$ for some constants $h$ and $g$. 
\end{proposition}
\noindent The full proof is in~\cite{LIU2020108758} where all constants are explicitly constructed. Conceptually, the theorem relaxes the standard Lyapunov conditions to allow a set $\Omega$ that contains the violation states where the Lie derivative can be positive ($\forall x\in \Omega, \lie V(x)\geq -aV(x)$). As long as each component of $\Omega$ is small enough (with volume less than $\varepsilon$), the violations do not affect stability. Under mild conditions, the appropriate region between sublevel sets of the system (written as $B$ in the definition) defines a forward invariant set for all trajectories, and an approximate form of contraction where the trajectory converges to near the lower level set of the region $B$. 

\begin{figure}[th!]    
    \centering
    \includegraphics[width=1\columnwidth]{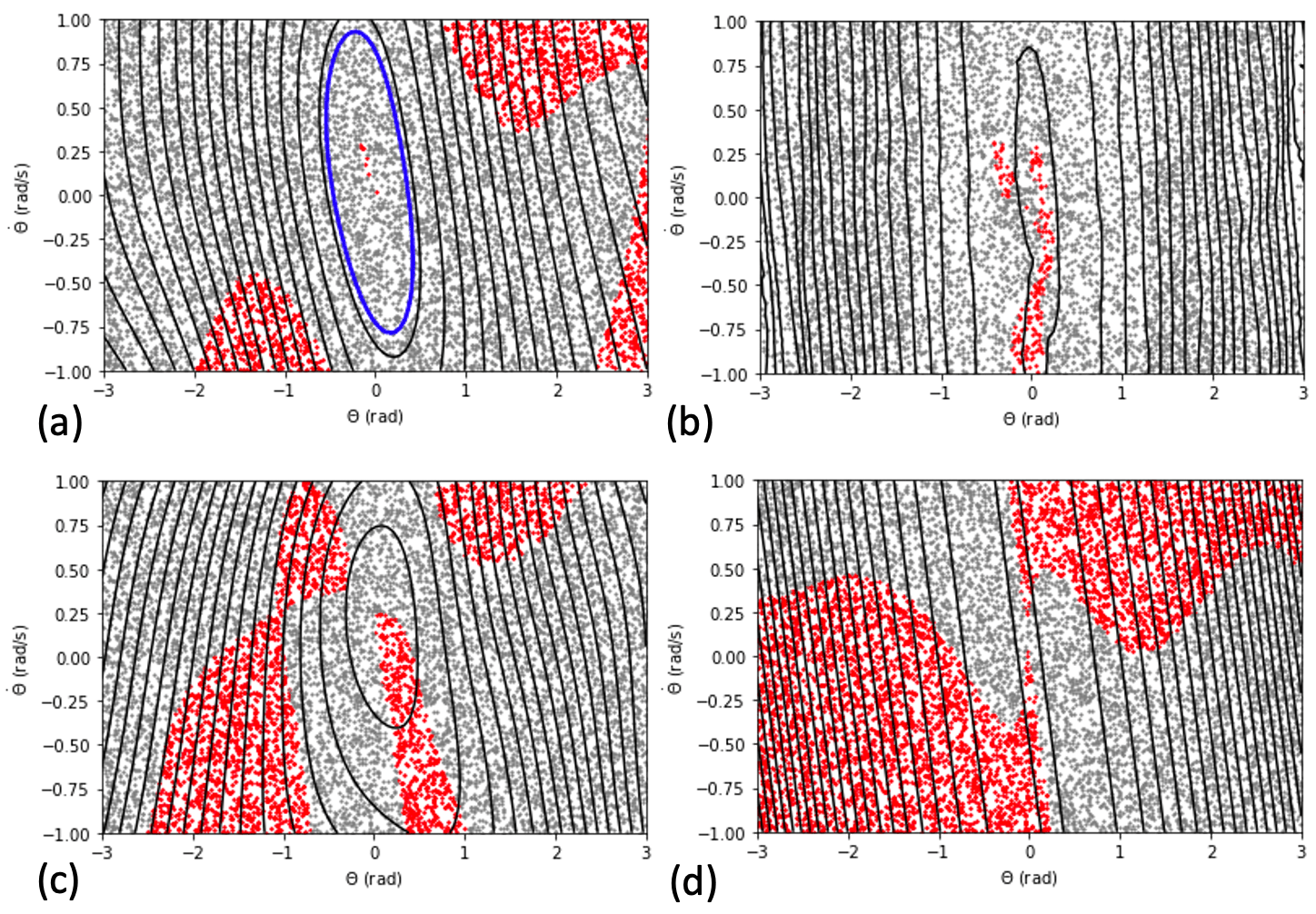}
    \caption{The landscapes of four different Lyapunov candidates for the inverted pendulum controlled by neural network policies.}\vspace{-.2cm}
    \label{fig:barrier}
\end{figure}

With the Almost Lyapunov conditions, we can use an $\varepsilon$-net over the space, chosen based on the Lipschitz constant of the Lie derivative, such that using sampled $\dlie V$ value at the center of each cell we can identify the set $\Omega$ of states where the Lie derivatives violate the standard conditions. In Figure~\ref{fig:barrier}, we show the results of such computation to compare four different types of candidate Lyapunov functions for the inverted pendulum controlled by neural network policies. Each plot visualizes the sign of the Lie derivative of the proposed Lyapunov candidate at uniformly sampled points over the $\theta/\dot\theta$ space. The grey dots represent cells where the candidate Lyapunov function has negative Lie derivative values ($\lie V<-aV(x)$ as required in Proposition~\ref{prop:lyp}), and the red dots indicate cells that violate such condition. The value of the Lyapunov candidate itself can be shown to be always nonnegative in the domain in all four cases, and the black contours represent the level sets of increasingly positive values. The red dots indicate states where the standard Lyapunov conditions are violated, and the patterns are different. 

$\bullet$ Figure~\ref{fig:barrier}(a) shows the self-learned Lyapunov critic obtained after learning the neural control policy using Algorithm 1. We see that we can find a sublevel set of the Lyapunov critic (inside the blue circle) where the Lie derivative is positive only at a very small number of sparse regions (a few red dots near the center). This landscape satisfies the Almost Lyapunov conditions and the sublevel set defines forward invariant set with attraction (Proposition~\ref{prop:lyp}). 

$\bullet$ Figure~\ref{fig:barrier}(b) shows the landscape generated by the Lyapunov actor-critic method (LAC)~\cite{DBLP:journals/corr/abs-2004-14288}. We see that the Lyapunov conditions are satisfied more globally, although with more violations closer to the origin. The function can also be established as an Almost Lyapunov function where the violation is sparse, which does not include the innermost sublevel set. The level sets is also much larger than those in (a), making it harder to find sublevel sets that are forward invariant. On the other hand, as shown later in the next section, the learned controller does always stabilize the system in all sampled trajectories. This indicates that there may exist better Lyapunov candidates for certifying the stability. 

$\bullet$ Figure~\ref{fig:barrier}(c) shows the Lyapunov candidate fit for the control policy learned by standard PPO. We see that much more violation states are observed and it does not allow us to find a region where Almost Lyapunov Conditions can be validated. This indicates that the lack of Lyapunov critic makes it hard to enforce stability properties. 

$\bullet$ Figure~\ref{fig:barrier}(d) uses the quadratic Lyapunov function for the linearized inverted pendulum obtained through LQR directly as a Lyapunov candidate for the neural controller trained by Algorithm 1, same as the one used in (a). We see that the simple Lyapunov function does not satisfy the Almost Lyapunov conditions and fails to capture the stability properties of the learned controller.

\section{Experiments}

We now show experimental results on the proposed methods for various nonlinear control problems. Our implementation follows Algorithm 1 (referred to as LY) and optionally uses an additional entropy term in the advantage estimates. We compare the performance with the closely related work of Lyapunov-based Actor-Critic (LAC)~\cite{DBLP:journals/corr/abs-2004-14288}, and also standard implementations of Soft Actor-Critic (SAC)~\cite{pmlr-v80-haarnoja18b} and PPO~\cite{ppo}. We use the following control environments: the inverted pendulum, quadrotor control, automobile path-tracking, and Mujoco Hopper and Walker. 
\begin{figure}[h!]    
    \centering
    \includegraphics[width=1\columnwidth]{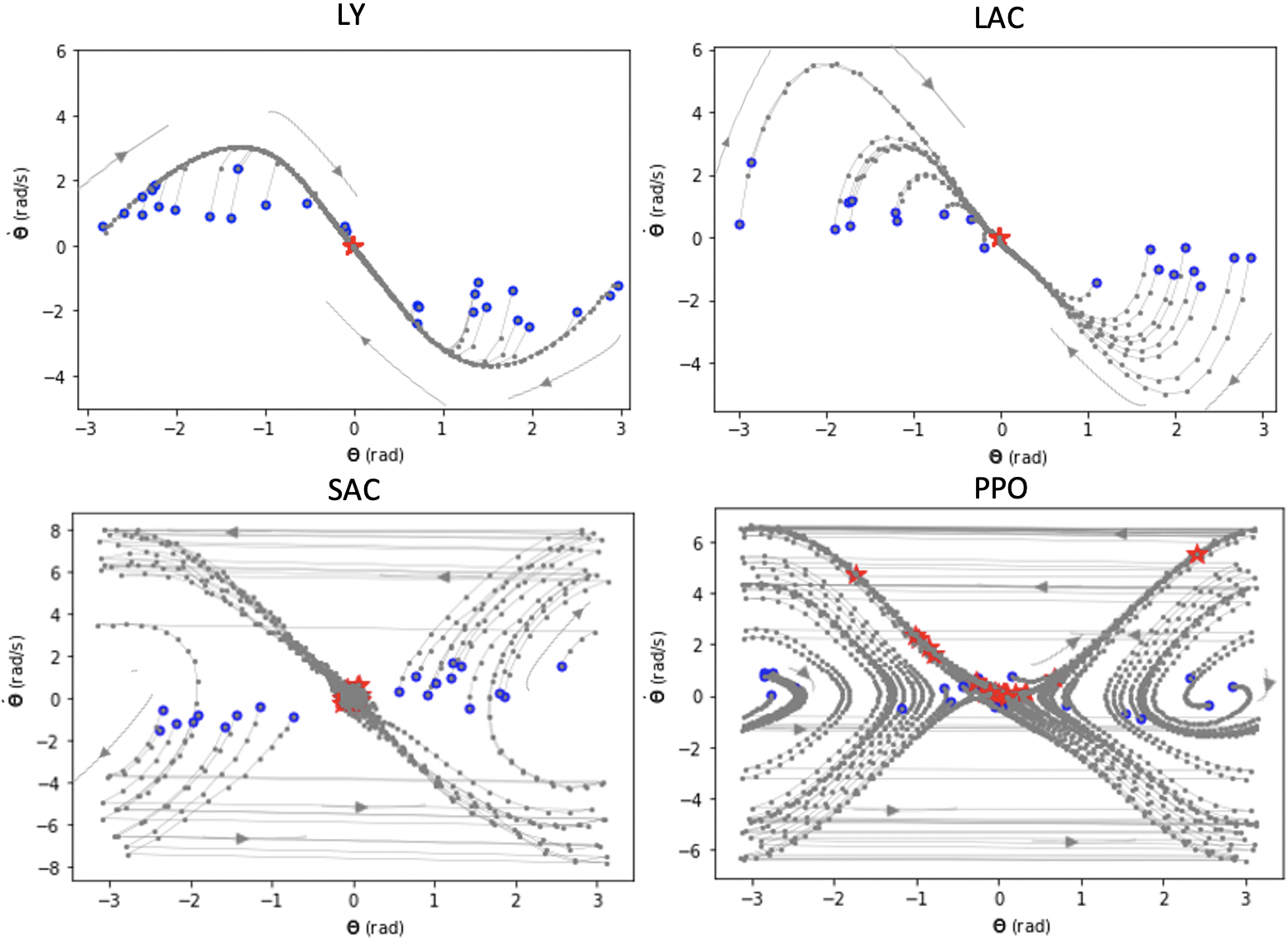}
    \caption{Sample trajectories generated by the policies learned with each algorithm in the inverted pendulum environment.}
    \label{fig:pendulum}
\end{figure}

\noindent{\bf Inverted Pendulum.} For the standard inverted pendulum (Pendulum-v0 in OpenAI Gym), Figure \ref{fig:pendulum} shows the system trajectories under learned policies plotted in the $\theta$-$\dot\theta$ space. The blue dots indicate initial positions and the red dots indicate positions after stabilization. Both LY and LAC learn stable controllers that reach the upright position without falling down, i.e., showing monotonically decreasing $\theta$ values. In contrast, the policies trained with SAC and PPO show horizontal lines that indicate falling down and crossing $-\pi$, by first swinging downwards and then back to the upright position. As discussed in the previous section in Fig 1, both LY and LAC learn Almost Lyapunov functions that are consistent with the stable control behaviors. 

\begin{figure}[th!]    
    \centering
    \includegraphics[width=0.9\columnwidth]{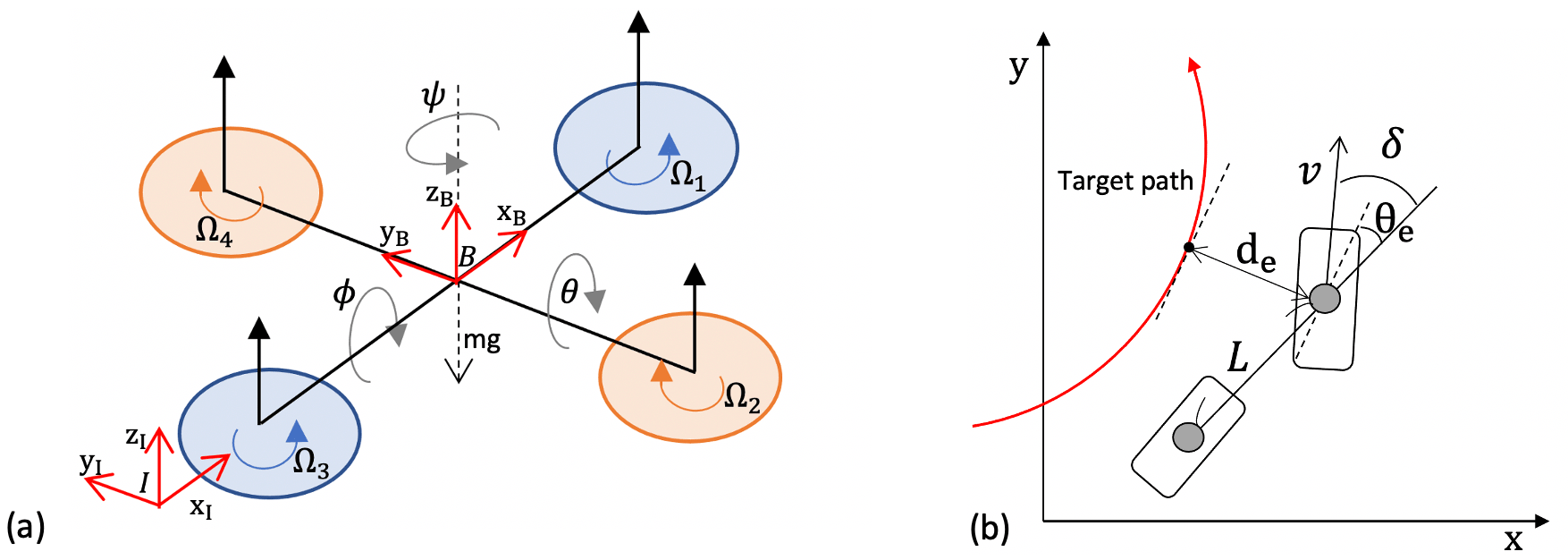}
    \caption{(a) Schematic of 6-DOF quadrotor system with body frame $B$ and inertia frame $I$. (b) Schematic of wheeled vehicle.}
    \label{fig:schematic}
\end{figure}

\begin{figure}[th!]    
    \centering
    \includegraphics[width=0.4\textwidth]{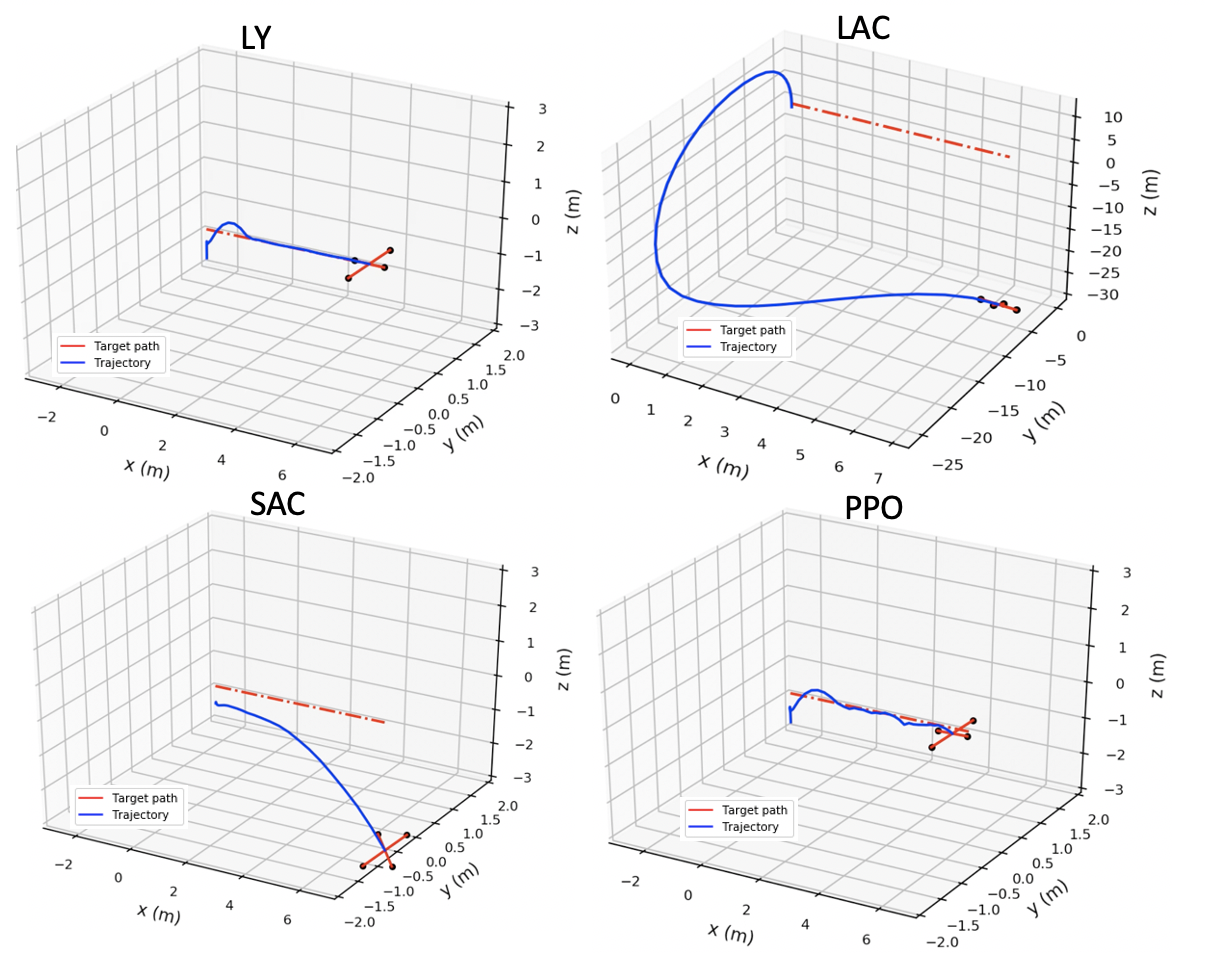}
    \includegraphics[width=0.8\columnwidth]{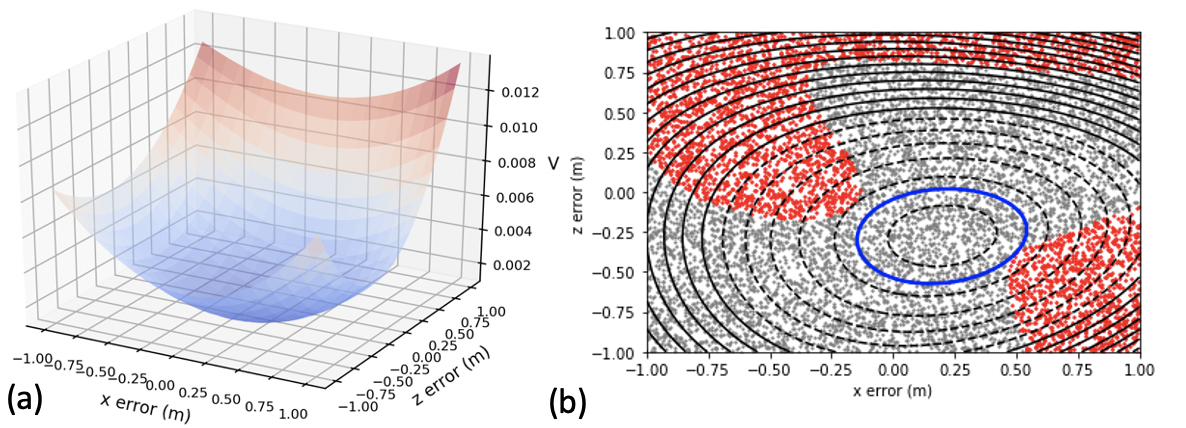}
    \caption{\small Quadrotor control for tracking a horizontal path. (a) shows the Lyapunov critic learned in LY and (b) shows where Almost Lyapunov conditions are validated (within the blue level set).}
    \label{fig:drone}
\end{figure}

\noindent{\bf Quadrotor control.} We learn neural controller for the 6-DOF  quadrotor model, which has four control inputs and twelve state variables. The control inputs $\Omega_{1}, \Omega_{2}, \Omega_{3}$ and $\Omega_{4}$ are the angular velocity of each rotor. The state variables are the inertia frame positions ($x, y, z$), velocities ($\dot{x}, \dot{y}, \dot{z}$), rotation angles ($\phi, \theta, \psi$) and angular velocities ($\dot{\phi}, \dot{\theta}, \dot{\psi}$). The equations of motion of the quadrotor are given in \cite{survey} and Figure \ref{fig:schematic}(a) shows the schematic. The control problem is known to be hard for policy gradient methods, typically requiring imitation learning steps. In Figure \ref{fig:drone}, we see that the LY method can learn stable tracking controller for the system. Moreover, the learned Lyapunov critic is shown in Fig 4(a) and its Lie derivatives is in Fig 4(b). Almost Lyapunov conditions can be validated with in the blue level set which certifies stable behavior shown in the trajectory. In comparison, LAC fails to learn a working controller. 

\begin{figure*}[t!h!]    
    \centering
    \includegraphics[width=0.9\textwidth]{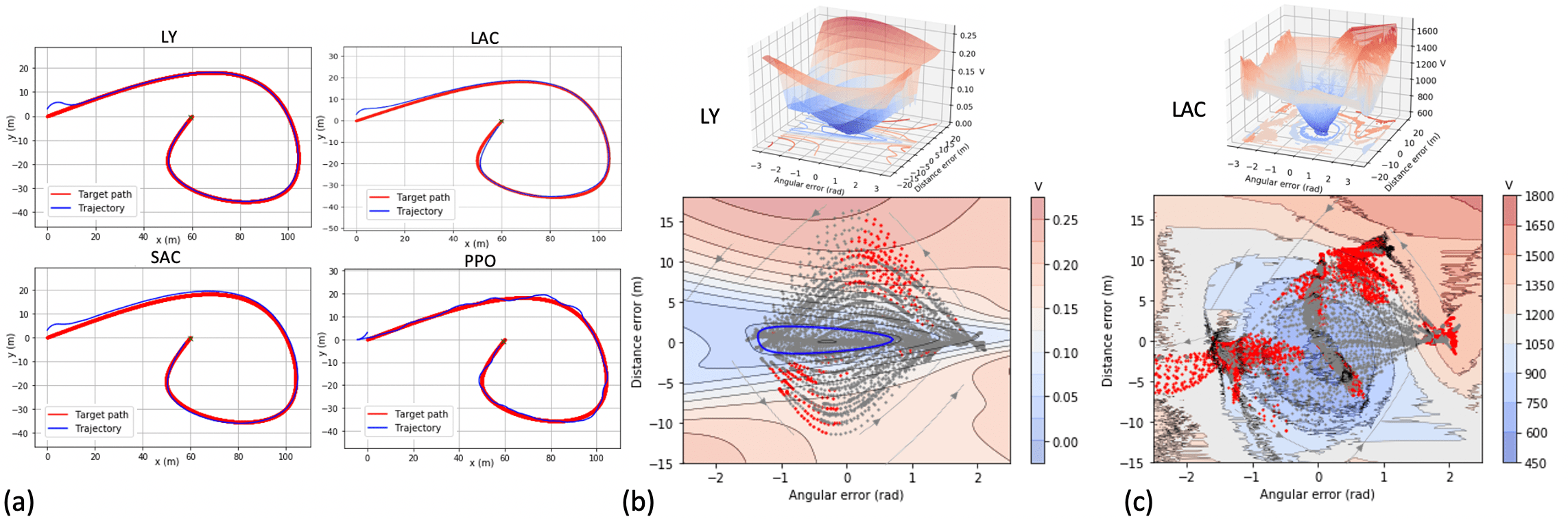}
    \includegraphics[width=1\textwidth]{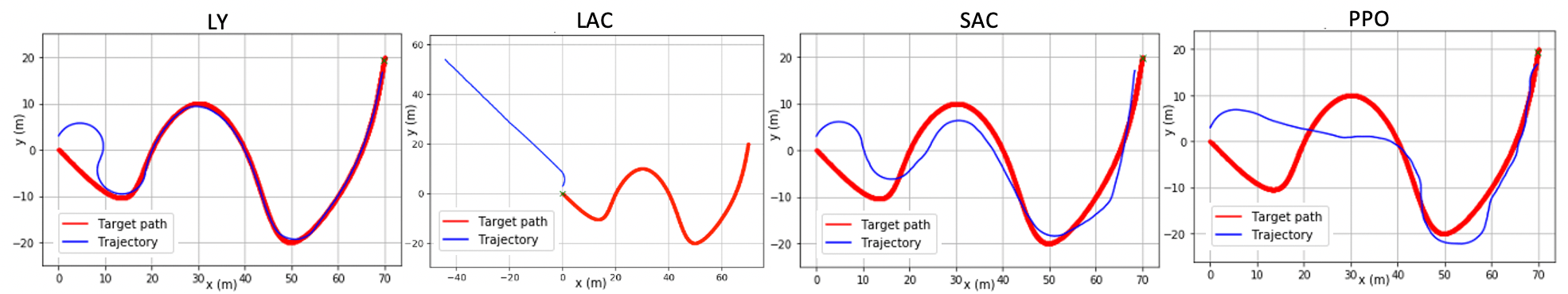}
    \caption{\small Experiments for automobile path tracking. (a) All methods learn to control well in the training environment. (b) 3D and 2D landscape generated by the learned Lyapunov function from LY. Almost Lyapunov conditions are validated within the blue level set. (c) Landscape generated by the learned Lyapunov critic from LAC. Second row: the control performance when tracking an unseen path (the red curve). The blue curves indicate the trajectory of the vehicle, starting from the left and going towards the right.}
    \label{fig:path}
\end{figure*}
\begin{figure*}[h!]    
    \centering
    \includegraphics[width=0.9\textwidth]{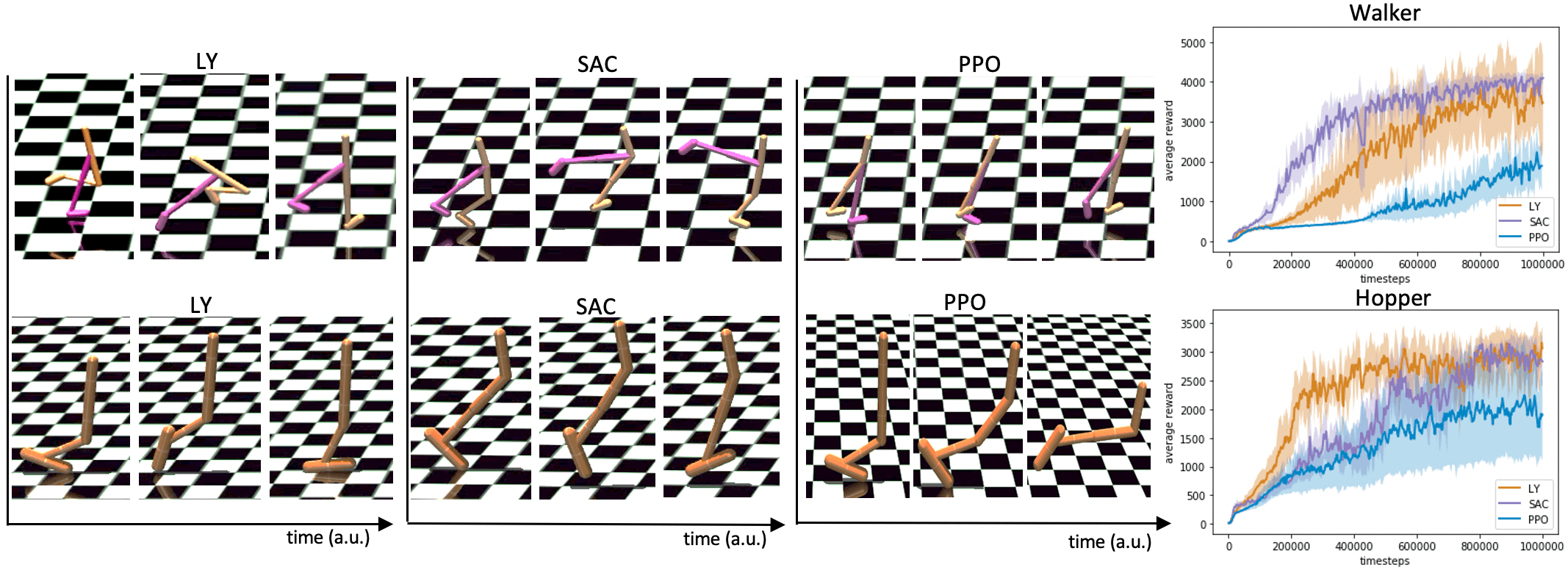}
    \caption{\small Control performance in the Walker and Hopper environments and learning curves over 5 random seeds. Our method typically learn faster than PPO, and comparable to SAC. All methods can achieve high rewards, but the learned control behaviors are different.}
    \label{fig:W&H}
\end{figure*}
\noindent{\bf Automobile path-tracking control.} The control problem of following a target path and speed command is the core of autonomous driving~\cite{article}. In the training environment, the state has four dimensions including target velocity $V_{t}$, angular error $\theta_{e}(t)$, distance to the path $d_{e}(t)$, and the vehicle speed $v(t)$. The action space contains acceleration $a(t)$ and a steering control $\delta(t)$. The car dynamics follows standard bicycle model~\cite{article} and Figure \ref{fig:schematic}(b) shows a schematic. 
In this experiment, we observe that all methods can obtain working control policies in the training environment in Figure \ref{fig:path}(a). However, the Lyapunov landscape matters when applying the policy on a different path, as illustrated in the bottom row in Fig~\ref{fig:path}. LY methods generalizes well to unseen paths because of its region of attraction, validated via the Almost Lyapunov conditions within the blue level set of the learned Lyapunov critic. 
In Figure \ref{fig:path}(c), we see that the Lyapunov critic learned in LAC does not create a landscape that enforces a region of attraction in the sense of Almost Lyapunov conditions. 

{\bf Mujoco Walker and Hopper.} Walker and Hopper are standard high-dimensional locomotion environment. 
For both the goal is move forward as fast as possible and not stabilization. The LAC method requires using cost functions for stabilization objectives only, and thus does not work in these environments. We can learn Lyapunov critics that are independent from the reward, just focusing on stabilizing the joint angles to hold the upright positions. As shown in Fig~\ref{fig:W&H}, the LY controller maintains better pose and gaits compared to SAC and PPO. The learning curves show that LY does not slow down learning, and can be used in generic high-dimensional control tasks to improve performance. 

\section{Conclusion}

We proposed new methods for training stable neural control policies using Lyapunov critic functions. We showed that the learned Lyapunov critics can be used to estimate regions of attraction for the controllers based on Almost Lyapunov conditions. We demonstrated the benefits of the proposed methods in various nonlinear control problems. Future work includes further improving sample complexity of the Lyapunov critic learning as well as the validation process for ensuring the Almost Lyapunov conditions. 

\newpage
\bibliographystyle{IEEEtran}
\bibliography{refs}

\end{document}